# *pCTFusion*: Point Convolution-Transformer Fusion with Semantic Aware Loss for Outdoor LiDAR Point Cloud Segmentation

[a]Abhishek Kuriyal, [a]Vaibhav Kumar*, [b]Bharat Lohani

abhishek19@iiserb.ac.in, *vaibhav@iiserb.ac.in, blohani@iitk.ac.in

[a]Department of Data Science and Engineering, Indian Institute of Science Education and Research Bhopal, India

[b]Department of Civil Engineering, Indian Institute of Technology Kanpur, India

*Corresponding author: Vaibhav Kumar

**Abstract**

LiDAR-generated point clouds are crucial for perceiving outdoor environments. The segmentation of point clouds is also essential for many applications. Previous research has focused on using self-attention and convolution (local attention) mechanisms individually in semantic segmentation architectures. However, there is limited work on combining the learned representations of these attention mechanisms to improve performance. Additionally, existing research that combines convolution with self-attention relies on global attention, which is not practical for processing large point clouds. To address these challenges, this study proposes a new architecture, *pCTFusion*, which combines kernel-based convolutions and self-attention mechanisms for better feature learning and capturing local and global dependencies in segmentation. The proposed architecture employs two types of self-attention mechanisms, local and global, based on the hierarchical positions of the encoder blocks. Furthermore, the existing loss functions do not consider the semantic and position-wise importance of the points, resulting in reduced accuracy, particularly at sharp class boundaries. To overcome this, the study models a novel attention-based loss function called Pointwise Geometric Anisotropy (PGA), which assigns weights based on the semantic distribution of points in a neighborhood. The proposed architecture is evaluated on SemanticKITTI outdoor dataset and showed a 5-7% improvement in performance compared to the state-of-the-art architectures. The results are particularly encouraging for minor classes, often misclassified due to class imbalance, lack of space, and neighbor-aware feature encoding. These developed methods can be leveraged for the segmentation of complex datasets and can drive real-world applications of LiDAR point cloud.

**Keywords:** Deep learning, LiDAR, Point cloud, Semantic segmentation, Attention mechanisms.

## 1 INTRODUCTION

Light Detection and Ranging (LiDAR) has become a critical component of development of many smart applications, primarily due to its ability of generating high fidelity 3D geometry of objects. The geometry information helps attain accurate 3D perception [1–4]. Consecutively, several advances have been made in semantic understanding of LiDAR point clouds using Machine Learning (ML) algorithms. Deep Learning (DL), a subset of ML, has displayed superior performance in point cloud processing capturing non-linear dependencies in complex scenarios [5].

Based on their underlying approaches, DL-based architectures can be categorized majorly into point-based [6], view-based [7], graph-based [8], and voxel-based [9] approaches. The three classes of algorithms, apart from point-based, construct an intermediate representation of the point cloud. View-based architectures train images generated by projecting point cloud on 2D plane, resulting in dimension and 3D information loss. Voxel-based models apply 3D kernels on voxel representation of the points. Although accurate, voxel-based approaches are computationally and memory-wise inefficient [10]. The graph-based approaches convert the point to subgraphs before any operation. The generation of optimal subgraphs is the key for this approach. Furthermore, due to changes in the object shapes, models result in less accuracies on sharp boundaries.

We extend the work on point-based architectures, which operate and rely on the geometrical distribution of point clouds without constructing any intermediate representations, e.g., voxels. To capture local geometric dependencies, point-based architectures use a variety of feature extractors such as attention mechanisms [10]. Some studies have devised novel strategies, e.g., positional encodings and attentive pooling [11], while some others have tried to model convolutions as native operations for point cloud. This includes grid-based networks such as PointCNN [6], graph-based networks such as DGCNN [12], kernel-based networks such as KPConv [13]. The reader can refer to the recent reviews by Guo et al.[14], Diab et al.[15], and Bello et al.[16] for more details related to various DL architectures and their application in classification, object detection, and segmentation of point clouds.

Unlike Multi-Layer Perceptrons (MLPs), Convolution Neural Network (CNN) based architectures impose a strong spatial inductive bias transferring local information to subsequent layers in neural networks. Recently, self-attention as an attention mechanism has been applied to point cloud processing [17, 18], delivering state-of-the-art performance in instance classification and part segmentation. Self-attention on local neighborhoods is more significant as point clouds display meaningful information within local regions particularly for large scale semantic segmentation tasks. Thus, recent attempts compute self-attention locally [19]. However, some approaches like M.-H. Guo et al.[17] and Lu et al.[20] utilize modified global attention on processed point clouds by downsampling number of points per cloud. This assists in reducing memory usage.

Despite the outstanding individual performance of the proposed mechanisms mentioned earlier, there has been limited exploration of their efficacy when employed in an ensemble. Recently, researchers have attempted to address this by integrating attention mechanisms, particularly transformer-based learning, and convolution for instance classification and part segmentation. Notably, Convolutional Point Transformer (CpT) and 3DCTN have been proposed by Kaul et al.[21] and Lu et al.[20], respectively. CpT leverages depth-wise convolution instead of linear projection to compute Query, Key and Value matrices, which are then passed through standard dot product attention. The output of the transformer layer and convolution layer are fused to incorporate both local and global contextual information. Meanwhile, 3DCTN utilizes both the coordinate context and feature context of neighboring points to generate combined point features, which are subsequently processed by 1x1 point-wise convolution kernels to compute edge features. The aggregated features obtained by max-pooling the above edge features are then passed through the offset attention which is a modification over vector self-attention to compute the final set of features. Major drawbacks of the above method are:

- Above-described methods employ global attention as the self-attention mechanism. However, this attention layer requires the entire point cloud to be consumed, resulting in an increase in sequence length for Query, Key, and Value matrices, which makes the process impractical for larger point clouds. To mitigate the computational overhead, these methods rely heavily on extensive downsampling.
- There is a lack of reported performance for semantic segmentation of outdoor point clouds, which are characterized by a large number of points. Our work is primarily motivated by the need to make this integration viable for outdoor point cloud segmentation, as it holds significant relevance for environment perception by autonomous systems.

While high performing encoders are crucial, an intricate loss function can prove equally beneficial. Such an attention mechanism can also be incorporated within the loss function to activate certain conditions during backpropagation. One major limitation of the previous point-based architectures is that they ignore the position and semantic importance of points. All the points are treated equal during backpropagation irrespective of their semantic and geometric distribution. However, the points along the separation boundaries are more prone to misclassification. To address this, we introduce " Pointwise Geometric Anisotropy Loss (*PGA Loss*)", a novel attention mechanism formulated as a loss function that assigns higher weights to the rich detailed neighborhoods in the point cloud.

**Contribution**

Our contribution is a novel architecture framework called Point Convolutional Transformer Fusion (*pCTFusion*), which combines kernel-based convolutions and self-attention based on the position of encoders in the network, along with a position-aware loss. The following are the key features of our proposed approach:

- Two encoder versions based on the position of encoders in the network. **Encoder V1**: This encoder version comprises an inception branch that integrates two kernel-based convolution units for multi-scale aggregation, followed by a local self-attention branch.

**Encoder V2**: Output of the inception branch (Encoder V1) is passed to the global self-attention branch, which is located as the last encoder in the network. To make global self-attention computationally efficient, the input point cloud is substantially downsampled at this encoder. The primary objective of Encoder V2 is to compute global dependencies at the final layer of the network.

- A novel loss function called "*PGA Loss*" This loss function assigns weights to the standard cross-entropy loss based on the distribution of semantic labels in the point cloud's neighborhood. Rather than providing equal semantic importance to each point, the *PGA Loss* assigns higher weights to areas in the point cloud that are more detailed and error prone.
- Exhaustive analysis highlighting the effect of each component of the network on the performance and the study of different parameters for each component.
- We also report results for a popular outdoor large scale point cloud dataset **SemanticKITTI** [22], and compare performance with state-of-the-art point-based architectures.

## 2 METHODOLOGY

Our proposed architecture framework, *pCTFusion*, attempts to combine two powerful attention mechanisms to obtain better feature representations while reducing unnecessary computational overhead thus striking a balance between performance and efficiency. The subsections describe the intricacies of our framework in detail. Table 1 summarizes the components of the architecture which are also the subsections of methodology section.

Table 1. Summary of the components (as subsections) of the architecture and their details

| Component | Description | Function |
|---|---|---|
| **Inception Branch** | Fusing 3D kernel-based convolutions for wide encoders. | Components of Encoder. |
| **Proposed Self-Attention Variant** | Computing dependencies between neighbourhood and globally. | Components of Encoder. |
| **Proposed Dual Encoder Version** | Integrating the Inception branch and self-attention creating dual encoders based on positions in the network. | Full Encoder Construction. |

| | | | |
|---|---|---|---|
| **Proposed *PGA Loss*** | | Incorporating attention mechanism within loss function to impart special attention to error prone regions. | Loss Function Construction. |
| **Proposed Architecture** | *pCTFusion* | The proposed architecture that integrates the previous subsections | The proposed architecture framework |

## 2.1 Inception branch for wide encoders

For a point cloud with N points, consider a representative point $p_i \in R^{(N,3)}$ with its feature $f_i \in R^{(N,f_{in})}$, the points and features of its neighbour can be described as $p_j \in R^{(N,3)}$ and $f_j \in R^{(N,f_{in})}$. The neighborhood relationship of points ($\Delta p_{ij}$) and features ($\Delta f_{ij}$) for the $i^{th}$ representative point can be explained as:

$$\Delta p_{ij} = p_j - p_i \quad (1)$$

$$\Delta f_{ij} = f_j - f_i \quad (2)$$

We choose a fixed disposition of polyhedron kernel k with a predetermined number of kernel points (see Figure 1). Each of these kernel $k$ has an associated weight assigned to it $w_k \in R^{(N_k,d_{in},d_{out})}$ where $N_k$ is the number of kernel points in each kernel, $(d_{in}, d_{out})$ is the feature associated with each kernel point.

The kernel weights cannot be directly multiplied to the neighbouring feature $f_j$ due to the lack of definite grid-like structure in the point cloud. To overcome this limitation, each neighbouring feature $f_j$ is multiplied to the weight assigned for every other kernel point along with a correlation function that produces higher value when Euclidean distance between $p_{ij}$ and the kernel points coordinates are small. Thus, the entire convolution operation for point $(p_i, f_i)$ is expressed by Equation (3).

$$f_{conv} = \sum_{k<N_k} \sum_{j<K} h(\Delta p_{ij}, p_k'') W_k \Delta f_{ij} \quad (3)$$

Here, $p_k'' \in R^{(N_k,3)}$ are the Euclidean coordinates of kernel point, $W_k \in R^{(N_k,d_{in},d_{out})}$ is the associated weight matrix with each kernel point and K is the number of nearest neighbours computed radially, $h$ is a linear correlation function chosen for the simplicity of backpropagation. The final step of convolution involves aggregating each of these neighbour features to the representative point. Above operation can be well described as an attentive aggregation of features in a neighbourhood with a tendency to introduce strong spatial inductive bias just like image convolution. One of the strengths of such a convolution framework is its ability to operate directly in the 3D metric space, thus having a richer and better capability to capture local dependencies. This capability is more crucial for scene understanding where global dependencies between each point in the cloud are not particularly important.

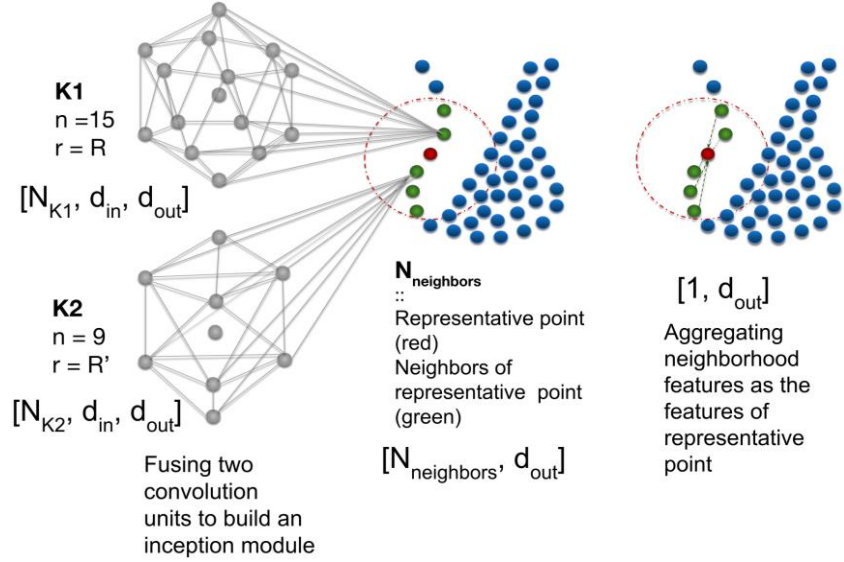

Figure 1: Inception module with two convolutions units for multi-scale feature aggregation

Rather than performing a single convolution, we fuse the feature representations of two such convolution operations with $(N_{k'}, R')$ and $(N_{k''}, R'')$ set of kernel parameters, respectively. Here, $N_{k'}$ and $N_{k''}$ are the number of kernel points while $R'$ and $R''$ are the influence radius of the kernels. The fusion captures the multi-scale feature representation and provides an overall balanced class-wise performance while the summation operator eases up backpropagation. We tuned and selected parameters via exhaustive experimentation. The fused convolution unit $f_{conv}$ can be described as:

$$f_{conv} = f'_{conv} + f''_{conv} \qquad (4)$$

## 2.2 Proposed Self-Attention Variants

The basic methodology of a self-attention operation is to convert the incoming feature vector into a pair of three vectors: Query (Q), Key (K), Value (V). These vectors provide an intuitive framework for the transformer model [23]. The standard scaled dot product attention $f_{global}$ (see Figure 2.a) formulated in the original work [23] computes global dependency between each query to every other key.

Considering an input vector of dimension $(N, f_{in})$ and output vector of dimension $(N, f_{out})$, $f_{global}$ can be formulated as:

$$f_{global} = \text{softmax}(\eta(Q \cdot K^T)) \cdot V \tag{5}$$

In our method, Q, K and V have dimensions $(N, f_{out}/4)$, $(N, f_{out}/4)$ and $(N, f_{out})$ respectively and η is the normalizing factor. For a large value of $N$, $f_{global}$ becomes computationally inefficient by increasing the memory footprint of the model. Thus, we only use $f_{global}$ in the last encoder where points are significantly down sampled.

Alternatively, self-attention can be reformulated to operate locally where dependencies are computed simultaneously in a neighbourhood. Contrary to natural language processing, computing local dependencies is more crucial for the point cloud semantic segmentation task. Hence, we first compute the radius neighbourhood for the incoming convolutional map using Equation 6.

$$f_* = neighbourhood(f_{conv}) \tag{6}$$

Here, $f_*$ has dimensions $(N, K, c_{in})$. The key components for self-attention query, key, and value matrices can be formulated using the below equations:

$$Q = W_Q \cdot f_{conv} \tag{7}$$
$$K = W_K \cdot f_* \tag{8}$$
$$V = W_V \cdot f_* \tag{9}$$

where $W_Q, W_K, W_V$ are the corresponding weights and $Q, K, V$ have $(N, 1, d_{out})$, $(N, K, d_{out})$, $(N, K, d_{out})$ dimensions, respectively. As shown in Equation 10, Local self-attention $f_{local}$ (see Figure 2.b), used in our approach, computes Hadamard product between Query and Key vectors operating in a neighbourhood.

$$f_{local} = \sum_{dim=1} softmax\left(\kappa(Q \cdot K + \gamma(p))\right) \cdot (V + \gamma(p)) \tag{10}$$

*where*, $p$ is the positional encoding and κ is a lightweight MLP.

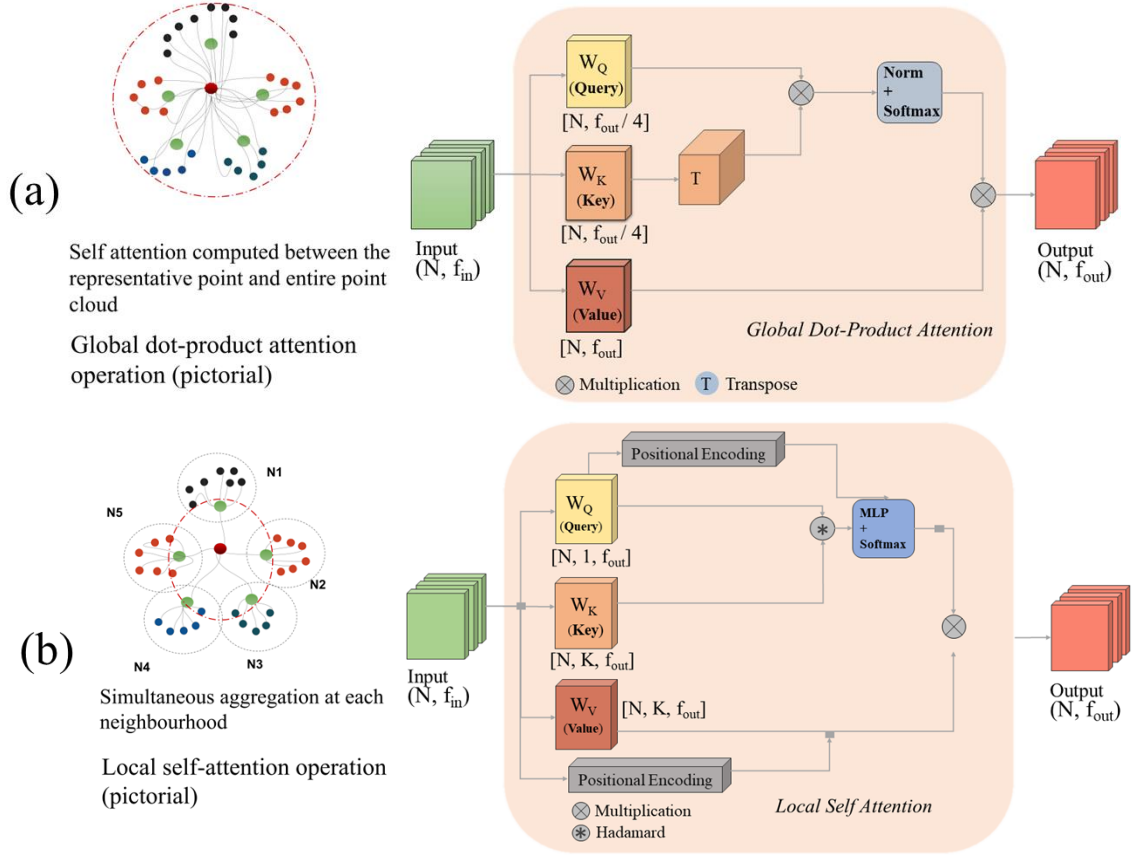

Figure 2: Difference between self-attention mechanisms for point cloud processing (a) global dot product (b) local self-attention.

Positional encoding has similar representation like Equation (1) and is used to supplement positional information in the local self-attention operation. The above formulation assigns SoftMax probability to each neighborhood in $V$ followed by a summation operation. Thus, the output from self-attention block ($f_{att}$) is computed based on the position of encoder in the architecture.

For last encoder (Encoder V2):

$$f_{att} = f_{global} \quad (11)$$

For rest others (Encoder V1):

$$f_{att} = f_{local} \quad (12)$$

### 2.3 Proposed Dual Encoder Versions

The obtained feature maps $f_{att}$ and $f_{conv}$ are then fused together via concatenation operation to leverage their individual strengths (see Equation 13).

$$f_{concat} = concat(f_{att}, f_{conv}) \quad (13)$$

Where, $f_{concat} \in R^{N, f_{out}/2}$. Output of an encoder block is formulated as Equation 14.

$$f_{enc} = LR\left(BN(MaxPool(f_{concat}))\right) \quad (14)$$

Where, $f_{enc} \in R^{N, f_{out}/4}$, $LR$, $BN$, $MaxPool$ are LeakyReLU, BatchNormalization and 1D MaxPooling operations, respectively. The above encoded features are passed through a shared MLP, which is formulated using the Equation 15.

$$f_{mlp} = LR\left(BN(Linear(f_{enc}))\right) \quad (15)$$

Here, $f_{mlp} \in R^{N, f_{out}}$ is connected to incoming features in a ResNet [24] fashion (see Figure 6). We designed two different versions of encoder depending upon how $f_{att}$ is computed. For the last encoder block where point density is very low, scaled dot product attention prove meaningful for the network to capture complex and long-range dependencies. Even though the employed local vector self-attention is computationally efficient, a lot of computation is incurred during indexing and gathering each point's neighbourhood. To mitigate it, we perform indexing operation only once for a particular encoding layer which is then used by each individual component in that layer. Figure 6 shows the two proposed encoder arrangements along with the overall network architecture.

### 2.4 Proposed Pointwise Geometric Anisotropy (*PGA Loss*)

Some segments of the input point cloud require special attention due to the semantic distribution with respect to their neighbouring points. Points located on the semantic separation boundaries, e.g., the feet of a human and the road, a pole and the ground etc., are more error prone. This is primarily because predicting the exact boundary line is a complicated task for the neural network (see Figure 3 for a few such boundary examples).

To mitigate such errors, we propose a novel Pointwise Geometric Anisotropy (**PGA**) attention mechanism inspired by Li et al. [25] formulated as a loss function for the network. The study computes PA-Loss based on the spatial importance of voxels while our method operates directly on points without any intermediate representation. The methodology of *PGA Loss* is as follows:

For a particular point $p$ having semantic label $l$ and its neighbor $p_j$ having semantic label $l_j$, PGA can be modelled as a $XOR$ operation between semantic labels of both points. Hence, $l\ xor\ l_j = 0$ if both the labels are same, otherwise $l\ xor\ l_j = 1$. Mathematically, for a fixed number of neighbours K, PGA between point $p$ and all its neighbours can be described using Equation 16.

$$PGA = \sum_{j<K} l\ xor\ l_j \tag{16}$$

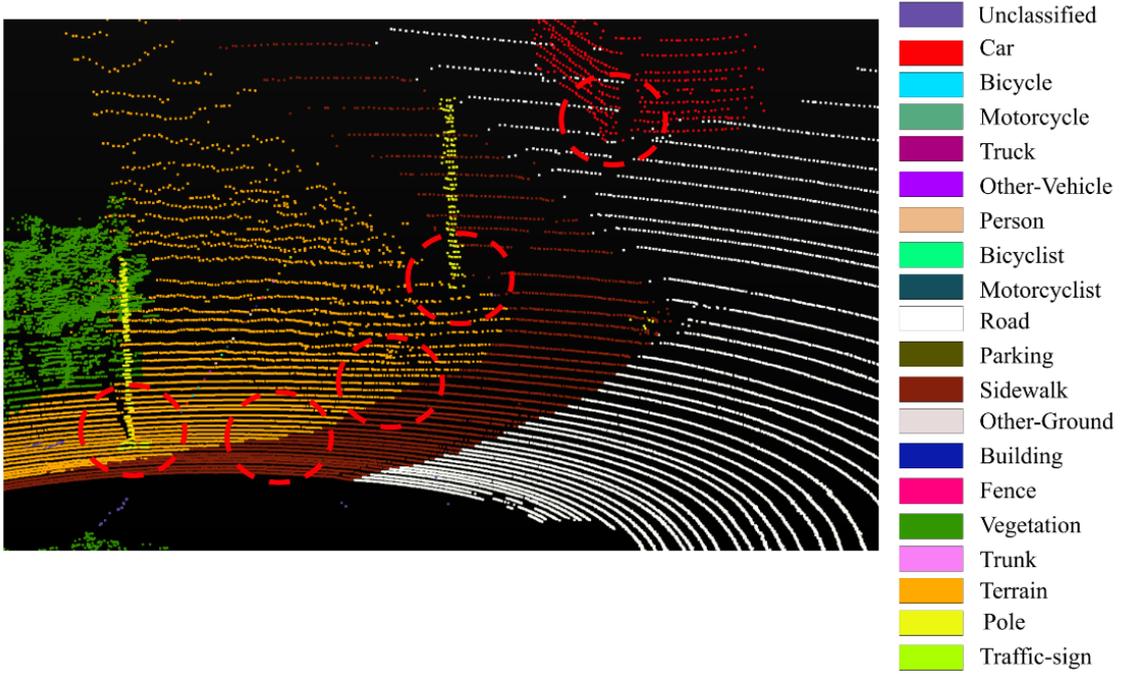

Figure 3. Instances of error prone regions located along the separation boundary between two classes in a point cloud.

Figure 4 (a) represents different scenarios of *PGA* score assignment based on the semantic labels between the representative point and the neighbouring points. Figure 4 (b) provides a pictorial representation of how *PGA* score is assigned for each point in the environment which is calculated using formulation in Equation (16). A high build-up along separation boundaries can be noticed in the figure. These points are given special attention during backpropagation as any misclassification would generate a high loss compared to the points with a low *PGA* score. Based on the *PGA* score for each input point, spatial importance score $W_{pga}$ can be defined by Equation (17), and this score is used to weight the standard cross-entropy loss using Equation (18).

$$W_{pga} = \eta + \theta \cdot PGA \tag{17}$$

$$L_{PGA} = \frac{-1}{N} \sum_{i=1}^{N} \sum_{c=1}^{n_c} W_{pga} y_i^t \log y_{ic}^p \qquad (18)$$

Here, $y_{ic}^t$ is the true label and $y_{ic}^p$ corresponds to softmax probability for $c^{th}$ class. Figure 5 shows representation of *PGA* Entropy assignment based on the neighbourhood interaction of the points.

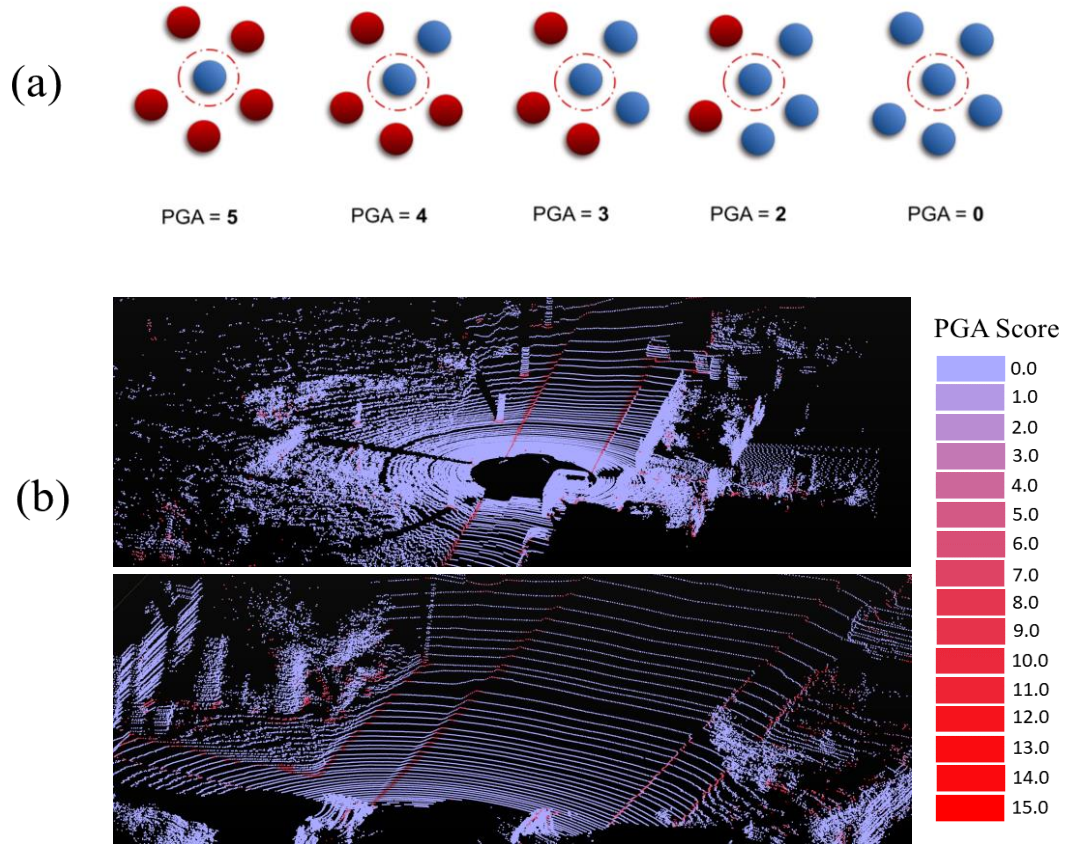

Figure 4: (a) Illustrates the PGA score assignment for two different classes highlighted with red and blue. Point enclosed within red boundary is the representative point. (b) PGA score assignment along different regions in the point cloud environment.

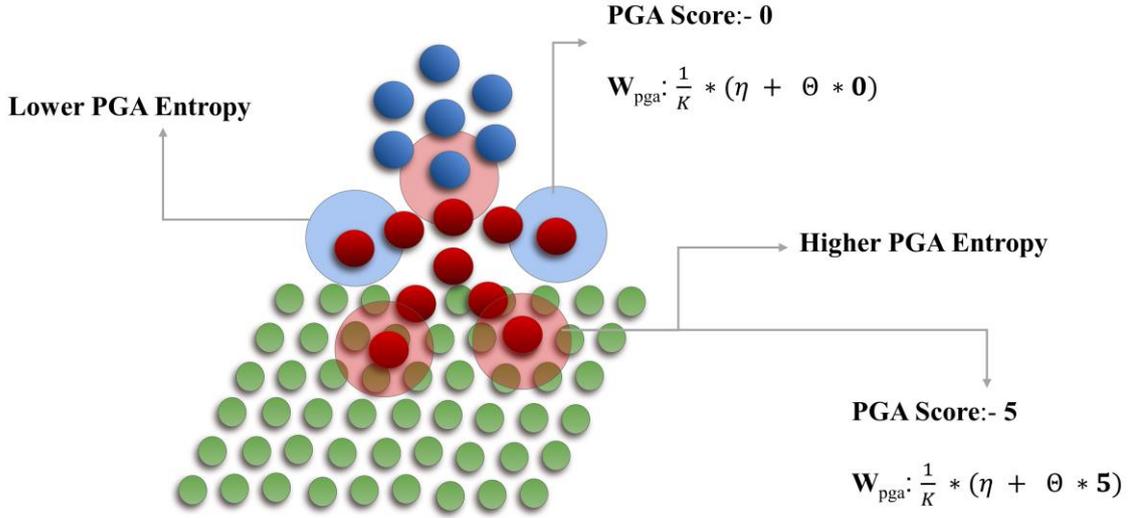

Figure 5: Pictorial representation of PGA entropy assignment based on the neighbourhood interaction of points. Blue corresponds to lower entropy region with lower PGA score while Red corresponds to higher entropy region with higher PGA score.

Considering a situation where points are located on the separation boundary, any misclassification would generate high $L_{pga}$ loss forcing the neural network to provide extra attention to these points. Theoretically, $L_{pga}$ can be computed after every decoding layer by upsampling the intermediate representation produced by the decoder to the original input dimension. This provides spatial attention to each decoding layer. However, in practice this will lead to too many loss functions for the backpropagation to minimize. Thus, we calculate $L_{pga}$ loss only for the last decoder output.

## 2.5 *pCTFusion*: Proposed Architecture

The proposed architecture integrates the methods discussed in the previous sections and consists of multiple stacked layers of encoder and decoder. Each encoder layer downsamples the point cloud while simultaneously increasing the per-point dimension and feature information. The decoder layer performs nearest neighbor upsampling progressively to obtain the final point-wise features. After the last decoding layer, the features are passed through fully connected Softmax layers to obtain a vector of dimension $(N, n_c)$. The *PGA Loss* $L_{pga}$ is computed for only the last decoder layer, but the same can be obtained for any intermediate decoding layer using a multi-step upsampling process. The convolution radius for each encoding layer is increased progressively to enhance the receptive field in each subsequent layer. Figure 6 provides an overview of the overall network architecture.

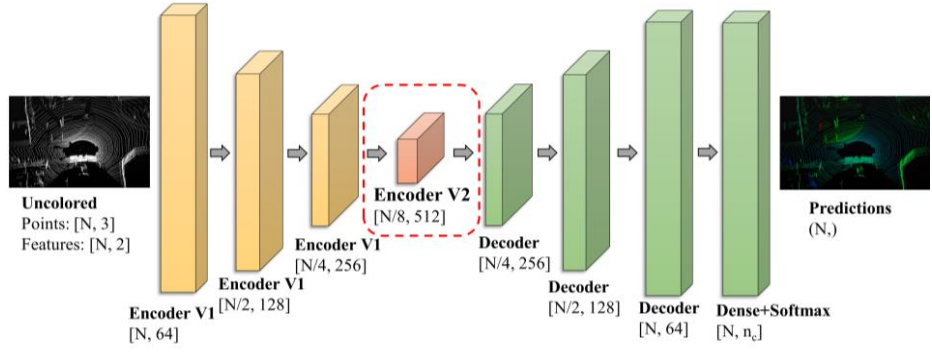

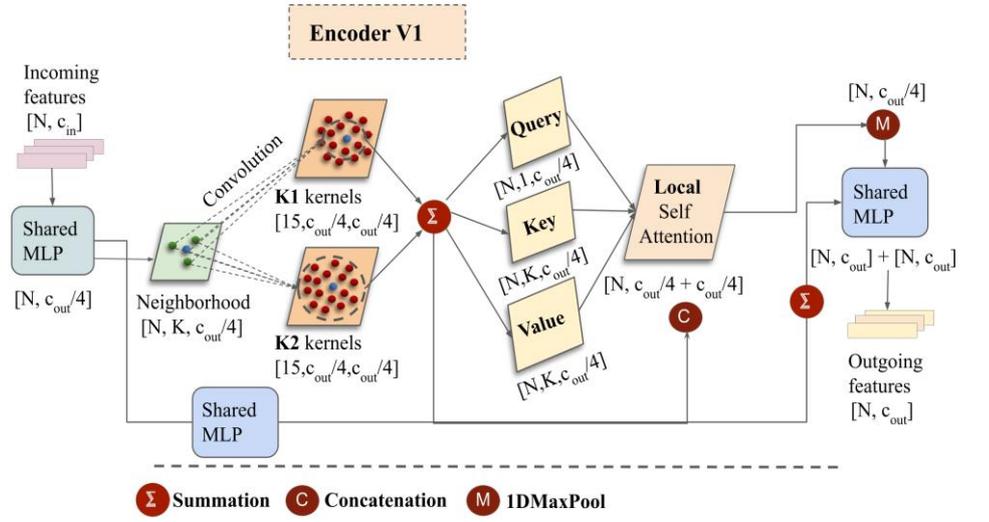

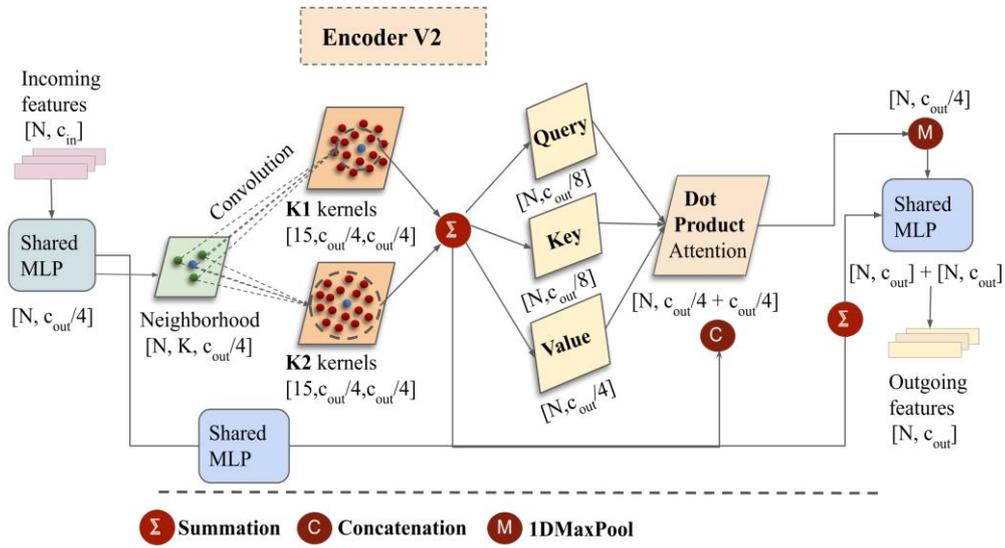

Figure 6: Overall network architecture of *pCTFusion* along with two proposed encoders architectures (Encoder V1 and Encoder V2). [26] Encoder V1 uses vector self-attention by operating on local neighborhood with K-nearest neighbors. Encoder V2 uses standard global dot product attention by operating on the entire point cloud. Encoder V2 is computationally feasible as the last encoder layer has low point density.

# 3 EXPERIMENTATION AND ANALYSIS OF OUTCOMES

We evaluated our proposed framework on a popular per-point segmentation dataset, i.e., SemanticKITTI [22]. The datasets contain different set of classes, hence, help in unbiased performance analysis of the architecture. Each dataset contains manually labelled points along with their cartesian coordinates and attributes such as intensity. After presenting the implementation outcomes the section we also discuss an ablation study on each component to highlight their contribution to the architecture's performance. We also compare the computational efficiency of our proposed method with the other state-of-the-art point-based architectures.

## 3.1 Implementation Details

In order to handle the large number of points in each individual point cloud in the datasets, a technique was employed where each scene was divided into small spheres. During training, each sphere was randomly selected and regularly tested using a voting scheme, with the final predicted result being the average of the predicted probabilities [13]. The SemanticKITTI dataset includes intensity information, which was combined with the range to create a feature vector. The neighbourhood computation was done radially instead of using K-nearest neighbours, as this approach was found to be more robust for varying densities [27]. Voxel subsampling was preferred over random subsampling, as it preserved the spatial structure of minor and smaller scale artifacts in the point cloud. The radius of convolution was set as 2.5 times the voxel subsampling radius (0.06), which was then doubled progressively for each encoding layer. These values were selected after trials and tests.

## 3.2 Dataset and Evaluation

The experiments were conducted using SemanticKITTI [22]. SemanticKITTI is a large-scale dataset consisting of point cloud scans of outdoor environments captured using Velodyne HDL-64E LiDAR mounted on car tops in the streets of Germany. This dataset is part of the KITTI Vision Odometry Benchmark and contains a total of 22 sequences, where sequences 00 to 10 are used as training sets and sequence 08 is used as the validation set.

**Metrics:** For evaluation, we compute Jaccard Index/Mean-Intersection-Over-Union ($mIoU$) [22]. Intersection over Union ($IoU_i$) for a class $i$ is calculated using Equation (18).

$$IoU_i = \frac{TP_i}{(TP_i + FP_i + FN_i)} \tag{18}$$

where $TP_i, FP_i, FN_i$ are the true positive, false positive and false negative scores for the class $i$. $mIoU$ is computed as the mean of $IoU$ for all classes.

## 3.3 Results on SemanticKITTI

This section discusses the results of our proposed architecture on sequence 08 (validation set) of the SemanticKITTI dataset. As shown in Table2, our architecture demonstrates superior overall performance compared to other well-known architectures. Furthermore, Figure7 provides a visual comparison between our model's predictions and the ground truth. It can be observed that our approach achieves excellent performance across various categories, including minor classes such as motorcycle, bicyclist, motorcyclist, pole, as well as major classes such as building, fence, vegetation, and more. These results further support the efficacy of our proposed methodology, which leverages attention and kernel-based fusion techniques for improved performance.

Table 2. mIoU score on sequence 08 (validation set) of SemanticKITTI. Here, **bold** represents the highest and underline represents the second highest score obtained by the architectures.

| Methods | RandLA Net [11] | Range-Net++ [28] | SqueezeSeg V3 [29] | Minkowski Net [30] | KPConv [13] | Salsa-Next [31] | pCTFusion (Ours) |
|---|---|---|---|---|---|---|---|
| Car | 92 | 89.4 | 87.1 | 95 | <u>95.9</u> | 90.5 | **97** |
| Bicycle | 8 | 26.5 | 34.3 | 23.9 | <u>43.3</u> | **44.6** | 39.3 |
| Motorcycle | 12.8 | 48.4 | 48.6 | 50.4 | <u>65.9</u> | 49.6 | **67.3** |
| Truck | <u>74.8</u> | 33.9 | 47.5 | 55.3 | 61.1 | **86.3** | 59.8 |
| Other-Vehicle | 46.7 | 26.7 | 47.1 | 45.9 | 51.3 | <u>54.6</u> | **59.9** |
| Person | 52.3 | 54.8 | 58.1 | 65.6 | 34.6 | **74** | <u>68.4</u> |
| Bicyclist | 46 | 69.4 | 53.8 | <u>82.2</u> | 52.3 | 81.4 | **90.4** |
| Motorcyclist | 0 | 0 | 0 | 0 | **0.2** | 0 | <u>0.15</u> |
| Road | 93.4 | 92.9 | **95.3** | <u>94.3</u> | 92.2 | 93.4 | 92.8 |
| Parking | 32.7 | 37 | <u>43.1</u> | **43.7** | 37.6 | 40.6 | 37.4 |
| Sidewalk | 73.4 | 69.9 | 78.2 | 76.4 | <u>79.5</u> | 69.1 | **80** |
| Other-Ground | 0.1 | 0 | 0.3 | 0 | **3.4** | 0 | <u>1.2</u> |
| Building | 84 | 83.4 | 78.9 | 87.9 | <u>90.5</u> | 84.6 | **90.7** |
| Fence | 43.5 | 51 | 53.2 | <u>57.6</u> | **65.6** | 53 | **65.6** |
| Vegetation | 83.7 | 83.3 | 82.3 | 87.4 | <u>88</u> | 83.6 | **89.1** |
| Trunk | 57.3 | 54 | 55.5 | 67.7 | **71.4** | 64.3 | <u>69.2</u> |
| Terrain | 73.1 | 68.1 | 70.4 | 71.5 | <u>73.9</u> | 64.2 | **76.9** |
| Pole | 48 | 49.8 | 46.3 | <u>63.5</u> | 60.9 | 54.4 | **65.2** |
| Traffic-Sign | 27.3 | 34 | 33.2 | **43.6** | <u>43.3</u> | 39.8 | 42 |
| *Mean mIoU* | 50 | 51.2 | 53.3 | 58.5 | 58.5 | <u>59.4</u> | **62.7** |

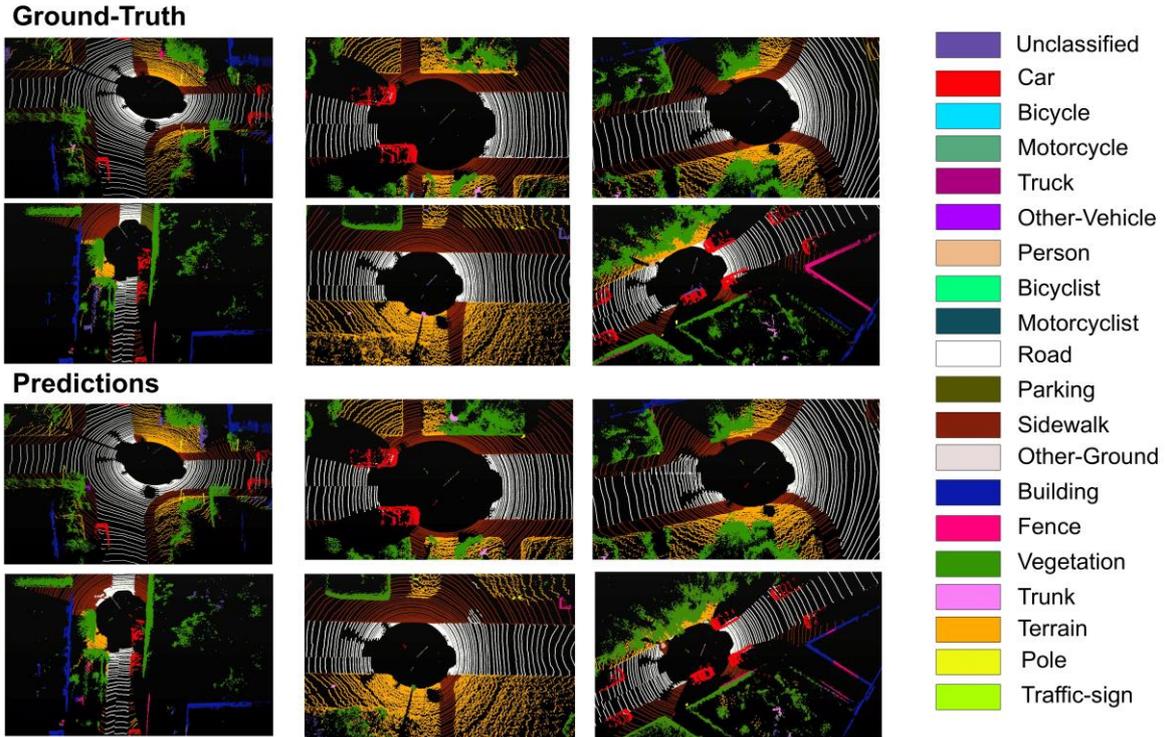
Figure 7: Visual comparison between ground truth and prediction results for SemanticKITTI validation set.

## 4 Ablation Studies

We conducted a detailed study of different governing components of our framework. The exhaustive analysis gives the details on *pCTFusion* components and their sensitivity on the segmentation performance. The analysis was carried out using SemanticKITTI dataset.

### 4.1 Qualitative analysis of *PGA Loss*

The *PGA Loss* enabled framework exhibits superior performance at sharp boundary regions of the point cloud, as depicted in Figure8 (circled in red). It accurately transitions and mitigates anomalies at separation boundaries to a significant extent. For instance, it successfully eliminates the anomaly of road class points that typically emerge at the transition boundary between sidewalk and terrain. Additionally, it assists the network in determining the degree of label spreading at these transition boundaries. This supports our hypothesis is that penalizing boundary areas compels the network to learn generalized features that effectively differentiate entities in these regions.

**Without PGA Loss enabled**

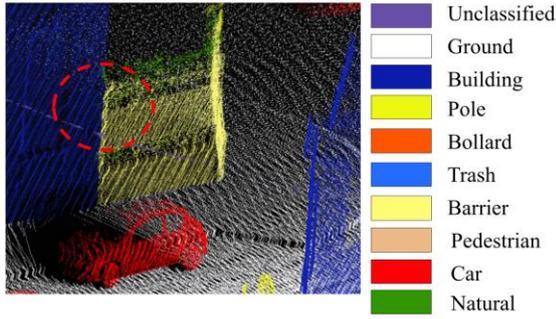
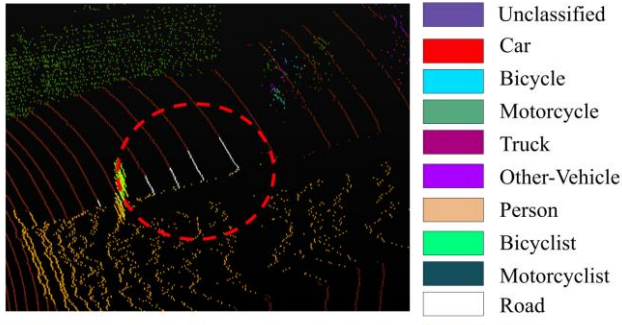

**With PGA Loss enabled**

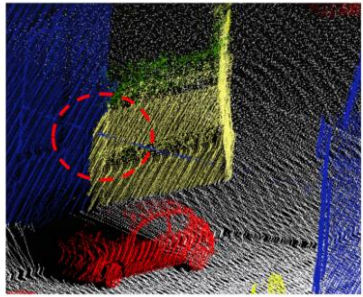
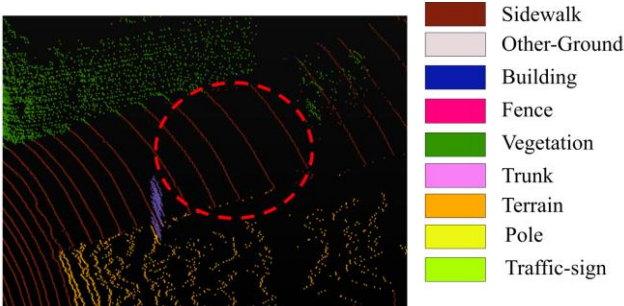

Figure 8: Visual comparison of PGA vs non-PGA-Loss enabled framework. The red circle highlights the area of interest where the PGA-enabled framework provides better boundary-level performance.

### 4.2 Analysing Individual Component of the Framework

Table 3 demonstrates the impact of each individual component of the network. Segmentation is done by employing each component in the ensemble one by one and mIoU is computed for each stage. We can observe a boost in performance by 2% after using two convolution units in the Inception module. The performance when using Encoder V2 version over the Encoder V1 version for the last encoding layer is subtle however the former can compute more complex dependencies for any generalized scenario due to presence of global self-attention between rich contextual information in last encoder layer. This helps in establishing a global relationship between points at last encoder where point density is low to provide greater information to each point other than just their respective neighbourhood The performance after using *PGA Loss* is increased by 1.73% demonstrating the special attention imparted by the same towards the error-prone areas in the point cloud. For higher density point clouds, the performance difference will be more substantial due to higher loss generation in such scenarios.

Table 3. Performance analysis of individual components used in our framework for SemanticKITTI data sample.

| 1st Convolution unit | 2nd Convolution unit | Encoder V1 as last encoder | Encoder V2 as last encoder | *PGA Loss* | mIoU % |
|---|---|---|---|---|---|

|  |  |  |  |  | 58.48 |
|---|---|---|---|---|---|
| ✓ |  |  |  |  | 58.48 |
| ✓ | ✓ |  |  |  | 60.42 |
| ✓ | ✓ | ✓ |  |  | 61.36 |
| ✓ | ✓ |  | ✓ |  | 61.53 |
| ✓ | ✓ |  | ✓ | ✓ | **62.73** |

### 4.3 Analysing Self-attention Types used in Encoder V1

We conducted a study of different plausible operations for the local self-attention component of Encoder V1 based on the class-wise performance for SemanticKITTI dataset. These operations are performed between Query (Q) and Key (K) vectors to establish dependencies within local neighbourhood [20]. We compared the impact of three major types of operations (see Table 4) on the model's performance: Summation, Subtraction and Hadamard (element-wise multiplication). We can observe comparative performance for both Subtraction and Addition operations with a minimal mIoU difference (~0.3) for SemanticKITTI.

We observed a significant performance boost using Hadamard (element-wise multiplication) operation for certain minor classes compared to subtraction and addition operations for e.g., bicycle (~3.95), other-vehicle (~6.75), person (~5.35), trunk (~6.35) etc. (here, numbers indicate an average performance increment). Similar performance increment can be observed with Paris-Lille 3D for certain minor classes like pole (~1.95), bollard (~3.30), trash-can (~4.55), barrier (~5.75), etc. In general, Hadamard operation proved more robust for both datasets by providing ~0.35 and ~1.80 percent of performance increment (averaged over all classes), hence, a natural choice in our formulation.

Table 4. Performance comparison of different local self-attention operation types used in our framework for SemanticKITTI 8[th] split.

| Operations | Subtraction (mIoU %) | Addition (mIoU %) | Hadamard (mIoU %) |
|---|---|---|---|
| Car | 95.1 | **96.4** | 96.3 |
| Bicycle | 39.3 | 38.2 | **42.7** |
| Motorcycle | **67.4** | 62.6 | 61.1 |
| Truck | **59.1** | 59.0 | 53.2 |
| Other-Vehicle | 54.3 | 49.2 | **58.5** |
| Person | 62.4 | 59.7 | **66.4** |
| Bicyclist | 77.3 | **78.2** | 72.7 |
| Motorcyclist | 0.2 | 0.16 | **0.2** |
| Road | **91.0** | 90.3 | 90.6 |
| Parking | **38.4** | 29.3 | 34.6 |
| Sidewalk | 76.1 | **83.1** | 80.4 |
| Other-Ground | 3.9 | 4.7 | **7.2** |
| Building | 92.7 | **94.5** | 92.2 |
| Fence | 61.1 | **62.3** | 62.0 |
| Vegetation | **85.5** | 82.1 | 82.4 |
| Trunk | 71.5 | 66.2 | **75.2** |
| Terrain | 72.5 | 71.7 | **72.7** |
| Pole | 60.9 | **66.5** | 61.3 |
| Traffic-Sign | **48.1** | 50.6 | 47.7 |
| *Mean mIoU* | 61.5 | 61.2 | **61.7** |

## 4.4 Analysing different Kernel Fusion Parameters in Inception Module

We explore different configurations and parameters for kernels involved in the fusion process. Overall, odd pairs of both K1 and K2 kernel sets showcased superior performance than the odd-even pair. A particular kernel configuration that performed best has the parameters ($N_k = 9, r = 0.75R$) for the K1 kernel set and ($N_k = 15, r = R$) for the K2 kernel set (~ 1.2% and ~ 1.4% mIoU difference for SemanticKITTI w.r.t to first entry in Table 5). This configuration resulted in better performance for some minor classes like motorcycle, bicyclist, pole, pedestrian etc. We believe this performance increment is most likely due to the lesser global energy hence stable geometry of the odd kernel arrangement comparative to even kernels. Since we desire to establish a close representative for 2D kernels using the above 3D kernel dispositions, a natural and stable geometry is most likely to perform better.

Table 5. Analysis of different kernel fusion parameters settings for SemanticKITTI. **Bold** text denotes the best outcome.

| Parameters | K1=10, K2=15, r=0.75R, r=R (mIoU %) | K1=15, K2=20, r=0.75R, r=R (mIoU %) | K1=9, K2=15, r=0.50R, r=R (mIoU %) | K1=15, K2=21, r=0.50R, r=R (mIoU %) |
|---|---|---|---|---|
| Car | 95.6 | 95.7 | **96.2** | 96.1 |
| Bicycle | 37.8 | 40 | 40.3 | **42.4** |
| Motorcycle | 62.4 | 64.2 | **66.4** | 64.3 |
| Truck | 54.9 | 51.9 | 61.1 | **61.4** |
| Other-Vehicle | 47.6 | **52.3** | **52.3** | 51.2 |
| Person | 61.5 | 63.9 | 65.5 | **66.3** |
| Bicyclist | 72.9 | 75.0 | **76.5** | 74.5 |
| Motorcyclist | 0.3 | 0.3 | **0.3** | 0.12 |
| Road | 92.2 | **92.3** | 92.1 | 90.3 |
| Parking | **39.6** | 37.4 | 36.4 | 32.4 |
| Sidewalk | **79.2** | **79.2** | 79.1 | 76.1 |
| Other-Ground | 3.1 | 5.4 | 5.9 | **6.1** |
| Building | **90.8** | **90.8** | 90.7 | 87.3 |
| Fence | **66.1** | 65.8 | 65.1 | 62.4 |
| Vegetation | 88.5 | 88.5 | 88.5 | 84.3 |
| Trunk | 69.6 | **70** | 69.5 | 68.4 |
| Terrain | 75.1 | 75.4 | 75.5 | **76.3** |
| Pole | 63.2 | **64.2** | 63.9 | 61.8 |
| Traffic-Sign | **46.5** | 45.8 | 46.1 | 44.3 |
| *Mean mIoU* | 60.4 | 60.9 | **61.6** | 61.4 |

## 5   Conclusion

This paper presented a novel approach for outdoor large-scale point cloud semantic segmentation by integrating attention mechanisms with convolution operations. Prior to this work, no research had attempted to combine these techniques for point cloud segmentation. To overcome this limitation, we proposed a novel architecture framework called Point Convolutional Transformer Fusion (*pCTFusion*), which combines kernel-based convolutions and self-attention-based mechanism to improve performance for per point segmentation tasks.

Our approach fused two different configurations of kernel parameters to leverage the strength of kernel-based convolutions. Additionally, we employed local and global self-attention mechanisms based on the hierarchical positions of the Encoder blocks to reduce computational overhead. The Encoder blocks computed indexed neighbours only once to further optimize performance. The architecture implemented proposed Pointwise Geometric Anisotropy Loss (*PGA Loss*) function that assigns weights based on the semantic distribution of points in a neighbourhood for assigning higher weights to the feature rich points. Such a loss function has not been implemented in the existing literature for point clouds and is one of the major contributions of the study.

The implementation of our proposed architecture yielded significant performance improvements compared to the state-of-the-art point-based architectures. Specifically, our approach achieved a 5-7% increase in performance when compared to MIL-Transformer, SIConv, and KPConv. Our approach proved particularly effective for minor classes and classes that are often misclassified by existing architectures due to their proximity to separation boundaries. Our architecture has the potential to be extended and implemented in complex environments with multi-sensor fusion settings. This would enable the classification of features with high accuracy, ultimately leading to better decision-making across various application domains. Overall, our proposed method has significant potential to enhance feature classification accuracy and improve decision-making in practical applications involving LiDAR point cloud.

**Codes**

The related codes can be accessed at: https://github.com/GeoAI-Research-Lab/PCTFusion.